\pdfoutput=1
\documentclass[11pt]{article}
\usepackage[final]{acl}
\usepackage{times}
\usepackage{latexsym}
\usepackage[T1]{fontenc}
\usepackage[utf8]{inputenc}
\usepackage{microtype}
\usepackage{inconsolata}

\usepackage{amsmath}  
\usepackage{amssymb}  
\usepackage{amsthm}   
\usepackage{graphicx}  
\usepackage{booktabs}  
\usepackage{relsize}
\usepackage{enumerate}
\usepackage{enumitem}
\usepackage{lipsum}
\usepackage{xspace}
\usepackage{cleveref}

\def\ie{\emph{i.e.}\xspace}

\title{Video-ChatGPT: Towards Detailed Video Understanding via \\
Large Vision and Language Models}

\author{%
Muhammad Maaz$^{1}$\textsuperscript{\textnormal{*}},
  Hanoona Rasheed$^{1}$\textsuperscript{\textnormal{*}}, 
  \textbf{Salman Khan}\textsuperscript{\textnormal{1,2}},
  \textbf{Fahad Shahbaz Khan}\textsuperscript{\textnormal{1,3}} \\
  \textsuperscript{1}Mohamed bin Zayed University of AI, UAE \\ 
  \textsuperscript{2}Australian National University, Australia \quad \textsuperscript{3}Linköping University, Sweden
    }

\begin{document}
\maketitle
\begin{abstract}
Conversation agents fueled by Large Language Models (LLMs) are providing a new way to interact with visual data. While there have been initial attempts for image-based conversation models, this work addresses the under-explored field of \emph{video-based conversation} by introducing Video-ChatGPT. It is a multimodal model that merges a video-adapted visual encoder with an LLM. The resulting model is capable of understanding and generating detailed conversations about videos. We introduce a new dataset of 100,000 video-instruction pairs used to train Video-ChatGPT acquired via manual and semi-automated pipeline that is easily scalable and robust to label noise. We also develop a quantitative evaluation framework for video-based dialogue models to objectively analyze the strengths and weaknesses of video-based dialogue models. Code: \url{https://github.com/mbzuai-oryx/Video-ChatGPT}.
\footnotetext[1]{Equal contribution.}
\end{abstract}

\section{Introduction}
The surge of deep learning applications for video understanding has lead to major advancements in video-related tasks. However, the current video understanding models are still unable to hold an open-ended conversation about the video content in a coherent manner. A video-based dialogue model can revolutionize video search, surveillance operations and help summarize key events and abnormal event detection. Above all, it can provide a unified human-understandable interface to video-related tasks such as action recognition, localization, detection, segmentation, retrieval, and tracking. Further, such a capability is of great interest as it will demonstrate the model's ability to encode temporal and spatial cues, contextual relationships and long-term dependencies. 

Recent advancements in multimodal understanding are largely based on the combination of pretrained \emph{image} models with Large Language Models (LLMs) but generally do not consider video inputs~\cite{liu2023llava, zhu2023minigpt, blip, blip-2, instructblip}. It is therefore interesting to leverage the vast capabilities of LLMs for video understanding tasks in a way that would not only maintain the temporal and spatial characteristics but also be adept at generating human-like conversations about videos. In this paper, we introduce Video-ChatGPT, a novel multimodal model that merges the representational abilities of a pretrained visual encoder and the generative powers of an LLM, capable of understanding and conversing about videos. 

Video-ChatGPT leverages an adapted LLM~\cite{liu2023llava} that integrates the visual encoder of CLIP~\cite{radford2021learning} with Vicuna~\cite{vicuna2023} as a language decoder, fine-tuned on generated instructional image-text pairs. Our approach further adapts the design for spatiotemporal video modeling and fine-tunes the model on video-instruction data to capture temporal dynamics and frame-to-frame consistency relationships available in video data. In contrast to other concurrent works for video-based conversation~\cite{2023videochat, damonlpsg2023videollama, su2023pandagpt}, Video-ChatGPT excels at temporal understanding, spatial consistency and contextual comprehension as demonstrated by our extensive evaluations. 

A fundamental contribution of this work is the creation of a dataset of 100,000 video-instruction pairs using a combination of human-assisted and semi-automatic annotation methods. Each pair consists of a video and its associated instruction in the form of a question-answer. This provides Video-ChatGPT with a large and diverse dataset to learn from, increasing its video-specific understanding, attention to temporal relationships and conversation capabilities.

Moreover, we introduce the first quantitative video conversation evaluation framework for benchmarking, allowing for a more accurate evaluation of the performance of video conversation models. This framework evaluates models on a variety of capabilities, such as correctness of information, detail orientation, contextual understanding, temporal understanding, and consistency. 

The contributions of this work are as follows,\vspace{-0.5em}
\begin{itemize}\setlength{\itemsep}{0mm}
    \item We propose Video-ChatGPT, a video conversation model capable of generating meaningful conversations about videos. It combines the capabilities of LLMs with a pretrained visual encoder adapted for spatiotemporal video representations.
    \item We introduce 100,000 high-quality video instruction pairs together with a novel annotation framework that is scalable and generates a diverse range of video-specific instruction sets.
    \item We develop the first quantitative video conversation evaluation framework for benchmarking video conversation models. We demonstrate Video-ChatGPT to perform well compared to concurrent conversational engines for videos such as Video Chat \cite{2023videochat}.
\end{itemize}

\section{Related Works}

\textbf{Vision Language Models:} Significant advancements in the field of computer vision have recently been observed due to the development of many foundational vision-language models. These models represent a significant leap towards creating general-purpose vision models capable of tackling various tasks simultaneously~\cite{radford2021learning, Alayrac2022Flamingo, gupta2022towards, Maaz2022Multimodal}. A prime example is CLIP~\cite{radford2021learning}, which is trained on 400M image-text pairs and has demonstrated impressive zero-shot performance on numerous benchmarks. It has been employed in various downstream applications, from image-based object detection and segmentation~\cite{Hanoona2022Bridging, liang2023open} to 3D applications~\cite{rozenberszki2022language, ni2022expanding}. Numerous attempts have also been made to adapt CLIP for video applications~\cite{wang2021actionclip, ni2022expanding}. Similar to our design, ViFi-CLIP~\cite{hanoonavificlip} suggests employing temporal pooling across video frames to adapt the image-based CLIP model for video-based tasks.

\noindent \textbf{Large Language Models:} The field of natural language processing has witnessed a paradigm shift with the advent of pretrained Large Language Models (LLMs) such as GPT~\cite{brown2020languagegpt}, LLaMA~\cite{touvron2023llama}, OPT~\cite{zhang2022opt}, and MOSS~\cite{OpenLMLab_2023}. These models exhibit extraordinary abilities like language generation and in-context learning, and their knack for understanding intricate tasks given user prompts in a zero-shot manner reflects their impressive adaptability and generalization. The proven capabilities of LLMs have encouraged researchers to fine-tune them to maximize their proficiency.

A key strategy in this pursuit is instruction tuning. This approach focuses on improving the model's alignment with user intentions and optimizing its output quality. For instance, InstructGPT~\cite{ouyang2022traininginstructgpt} and ChatGPT~\cite{chatgpt} significantly benefit from this technique, showcasing improvements in diverse conversational interaction capabilities and their aptitude to answer a broad range of complex questions. This effective approach has recently been employed in open-source models like Alpaca~\cite{alpaca} and Vicuna~\cite{vicuna2023}, both developed using the LLaMA~\cite{touvron2023llama} framework, resulting in performance improvements.

\noindent \textbf{Pre-trained LLMs in Vision-Language Tasks:} The recent strides in multimodal understanding have primarily been driven by the integration of image-based vision models with LLMs. Seminal contributions such as Flamingo~\cite{Alayrac2022Flamingo} and BLIP-2~\cite{blip-2} have demonstrated the power of utilizing web-scale image-text data, as well as pioneering techniques in cross-modal alignment, to exhibit dynamic abilities in conversational and few-shot learning contexts. Building on this foundation, MiniGPT-4~\cite{zhu2023minigpt} allows image-based conversations by integrating BLIP-2 and Vicuna for zero-shot image comprehension.

Equally significant is the emergence of LLaVA~\cite{liu2023llava}, a model derived from the LLaMa architecture, leveraging GPT-4's language proficiency to generate multimodal instruction-following data. With instruction tuning applied on the derived data, LLaVA has displayed interesting multimodal chat capability, hinting at the scalability potential of such a methodology. In addition, InstructBLIP~\cite{instructblip} has demonstrated strong image-based dialogue capabilities via vision-language instruction tuning by innovating with instruction-aware visual feature extraction.

More closely related to our work, VideoChat~\cite{2023videochat} employs selective components of video foundational models~\cite{wang2022internvideo} and image foundation models~\cite{blip-2}, and integrates them with LLMs ~\cite{vicuna2023} in conjunction with few learnable layers, tuned using a two-stage lightweight training. Additionally, they construct a video-specific dataset using off-the-shelf vision-language models~\cite{wu2022grit, blip-2, huang2023tag2text, wang2022internvideo} for generating noisy detailed textual descriptions to enhance the training of video-centric conversational models.

Different from VideoChat, we propose a novel human assisted and semi-automatic annotation framework for generating high quality instruction data for videos. Our simple and scalable architecture design utilizes pretrained CLIP~\cite{radford2021learning} to generate spatiotemporal features which help Video-ChatGPT in generating meaningful video conversation. Further, we are the first to propose quantitative framework for evaluating video conversation tasks (see Section~"Video Instruction Data Generation" for more details).

\begin{figure*}[t]
  \centering
  \includegraphics[width=0.99\textwidth]{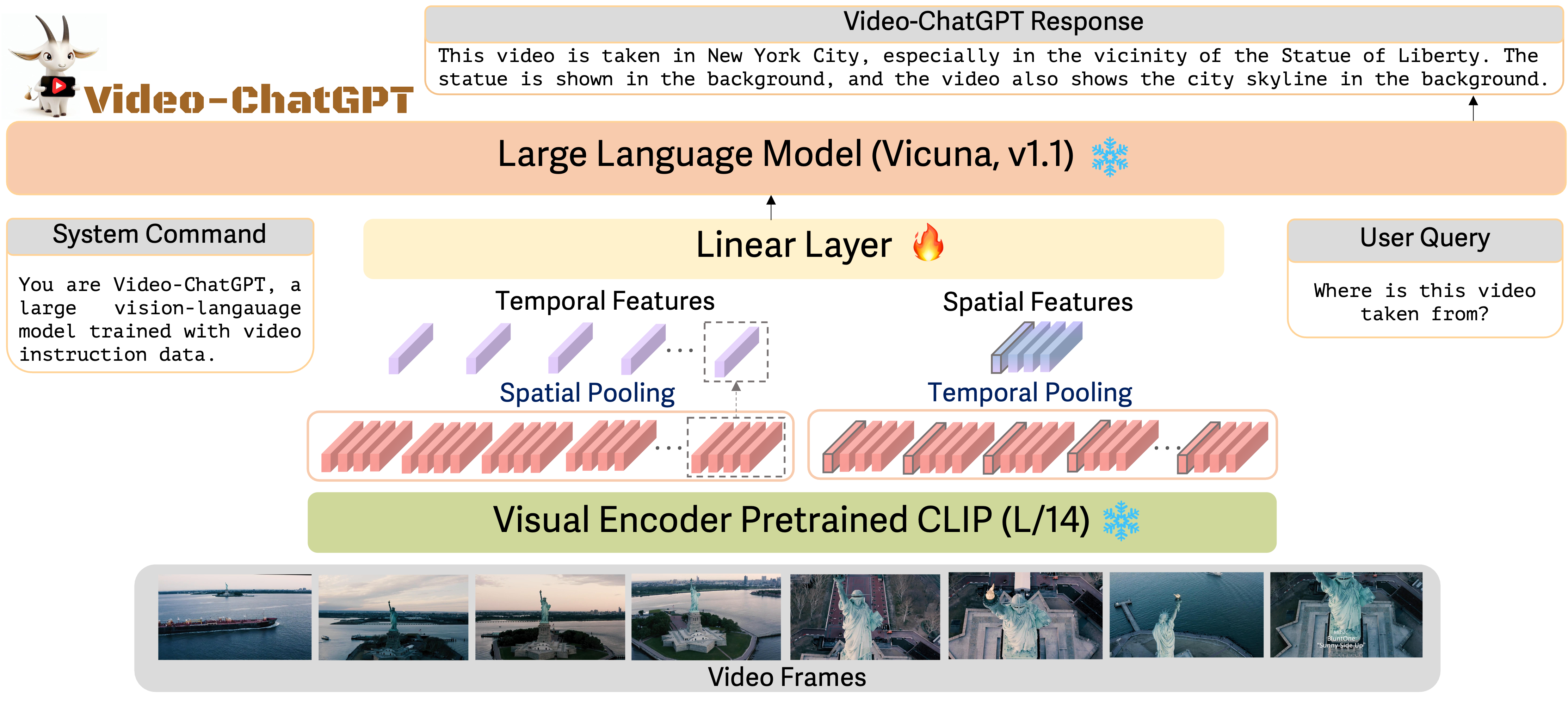}
  \caption{\textbf{Architecture of Video-ChatGPT.} Video-ChatGPT leverages the CLIP-L/14 visual encoder to extract both spatial and temporal video features. This is accomplished by averaging frame-level features across temporal and spatial dimensions respectively. The computed spatiotemporal features are then fed into a learnable linear layer, which projects them into the LLMs input space. In our approach, we utilize the Vicuna-v1.1 model, comprised of 7B parameters, and initialize it with weights from LLaVA~\cite{liu2023llava}.}
  \label{fig:arch}
\end{figure*}

\section{Video-ChatGPT}
Video-ChatGPT is a large vision-language model that aligns video representations with a Large Language Model (LLM), thus enhancing its ability to generate meaningful conversation about videos. Our approach draws from the approach employed in designing vision-language (VL) models for the video domain. Given the limited availability of video-caption pairs and the substantial resources required for training on such data from scratch, these models commonly adapt pretrained image-based VL models for video tasks~\cite{ni2022expanding, wang2021actionclip, hanoonavificlip}. We adopt a similar approach, starting with the Language-aligned Large Vision Assistant (LLaVA)\cite{liu2023llava} as our foundation.

LLaVA is a LMM that integrates the visual encoder of CLIP~\cite{radford2021learning} with the Vicuna language decoder~\cite{vicuna2023} and is fine-tuned end-to-end on generated instructional vision-language data. We fine-tune this model using our video-instruction data, adapting it for video conversation task. The video-instruction data is obtained as a combination of manual and automated pipelines in our proposed instruction generation setup. This adaptation on video-specific instructions allows for accommodating additional temporal dynamics, frame-to-frame consistency, and long-range relationships present in video data. As a result, our Video-ChatGPT excels in video reasoning, creativity, and understanding of spatial, temporal, and action-oriented components within videos.

\subsection{Architecture}
We use CLIP ViT-L/14, which is pretrained using large-scale visual instruction tuning in LLaVa, as the visual encoder. However, LLaVa visual encoder is meant for images, which we modify to capture spatiotemporal representations in videos. Given a video sample $V_i \in \mathbb{R}^{T \times H \times W  \times C}$ with $T$ frames, the visual encoder generates temporal and spatial features.
The visual encoder encodes the $T$ frames independently as a batch of images and produces frame-level embeddings $x_i \in \mathbb{R}^{T \times h \times w \times D}$, where $h=H/p, w=W/p$. Here p is the patch size (\ie 14 for ViT-L/14), and we represent the number of tokens as $N$, where $N = h \times w$. Frame-level embeddings are average-pooled along the spatial dimension to obtain a \textit{video-level temporal representation} $t_{i} \in \mathbb{R}^{T \times D}$. This operation implicitly incorporates temporal learning through the aggregation of multiple frames. Similarly, the frame-level embeddings are average-pooled along the temporal dimension to yield the \textit{video-level spatial representation} $z_{i} \in \mathbb{R}^{N \times D}$. The temporal and spatial features are concatenated to obtain the video-level features $v_{i}$, 
\begin{equation}
    v_{i} = [t_{i} \quad z_{i}] \in \mathbb{R}^{(T+N) \times D}.
\end{equation}
A simple trainable linear layer $g$, projects these video-level features into the language decoder's embedding space, transforming them into corresponding language embedding tokens $Q_{v}$,
\begin{equation}
    Q_{v} = g(v_{i}) \in \mathbb{R}^{(T+N) \times K} .
\end{equation}
Note that the function $g$ acts as an adapter and can be implemented with more complicated architectures as well. However, we opt for a simplistic design that gives competitive performance compared to more sophisticated choices in our experiments. 
The text queries are tokenized to the same dimensions, $Q_{t} \in \mathbb{R}^{L\times K}$. Here $L$ represents the length of text query.
Finally, $Q_{v}$ is concatenated with $Q_{t}$ and input to the language decoder.

\begin{figure*}[t]
  \centering
  \includegraphics[width=0.99\textwidth]{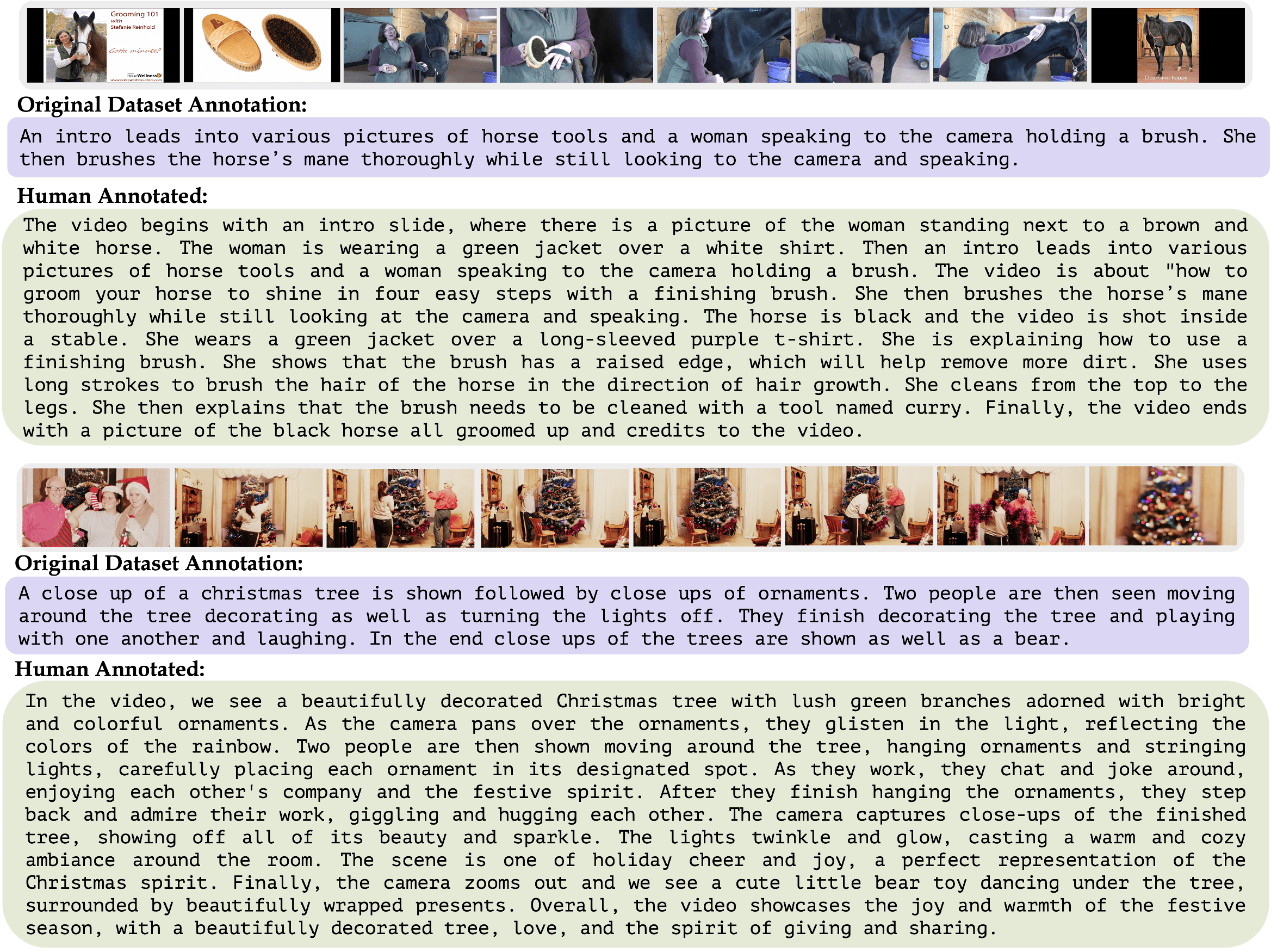}
  \caption{\textbf{Examples of data enrichment via human-assisted annotation}. Human annotators augment video descriptions from video-caption datasets. The captions are enriched by integrating detailed information about spatial and temporal aspects, object relationships, reasoning, scene descriptions, and the chronological sequence of events.}
  \label{fig:human_assisted}
\end{figure*}

\subsection{Video Instruction Tuning}
We employ instruction-tuning of the LLM on the prediction tokens, utilizing its original auto-regressive training objective. The pretrained model is finetuned with curated, high-quality video-text pairs. During the finetuning phase, we use predefined prompts based on the following template:
\begin{center}
\noindent\texttt{USER: <Instruction> <Vid-tokens> Assistant:}
\end{center}
\noindent Using the notations, we can represent it as,
\begin{center}
\noindent\texttt{USER: <$Q_{t}$> <$Q_{v}$> Assistant:}
\end{center}

In this prompt, the \texttt{<Instruction>} represents a question pertaining to the video, randomly sampled from the training set of video-question-answer pairs. Questions can be general, asking to describe the video, or they may relate to specific temporal, spatial, or creative aspects of the video content. The prediction answer \texttt{<Answer>} corresponds to the specific question asked. Throughout the training, the weights for both the video encoder and LLM remain frozen, and the model maximizes the likelihood of predicting tokens representing the answer by adapting the linear layer. Consequently, the video features $Q_{v}$ become aligned with the pre-trained LLM word embeddings, equipping Video-ChatGPT with the ability to produce more natural and dependable responses.

\section{Video Instruction Data Generation}
\label{video_instruction_data}
In this section, we discuss our data-focused approach, which uses both human-assisted and semi-automatic annotation methods to generate high-quality video instruction data. This data is crucial for training Video-ChatGPT, ensuring accurate and meaningful responses. Our data collection involves two key methods. The \textit{human-assisted annotation}, involves expert annotators analysing video content and providing detailed descriptions. This process generates data rich in context and detail, which helps our model understand complex aspects of video content.
On the other hand, the \textit{semi-automatic annotation framework} is more cost-effective and scalable. Leveraging state-of-the-art vision-language models, this method generates broad, high-volume annotations, thus increasing the quantity of data without compromising the quality substantially.
Through these combined methods, we have successfully accumulated a robust set of 100,000 video-instruction pairs. This extensive dataset is crucial in fine-tuning our model to comprehend video content effectively, integrating both spatial and temporal cues into its understanding.

Our instructional data is both diverse and comprehensive, incorporating a wide range of data types. These include detailed descriptions, summarizations, question-answer pairs, tasks that stimulate creativity or generation of new ideas, and conversational tasks. The data spans a broad spectrum of concepts, ranging from visual appearance and temporal relations to complex reasoning tasks and beyond, providing a diverse training ground for our model to learn from.

\begin{figure*}[t]
  \centering
  \includegraphics[width=0.99\textwidth]{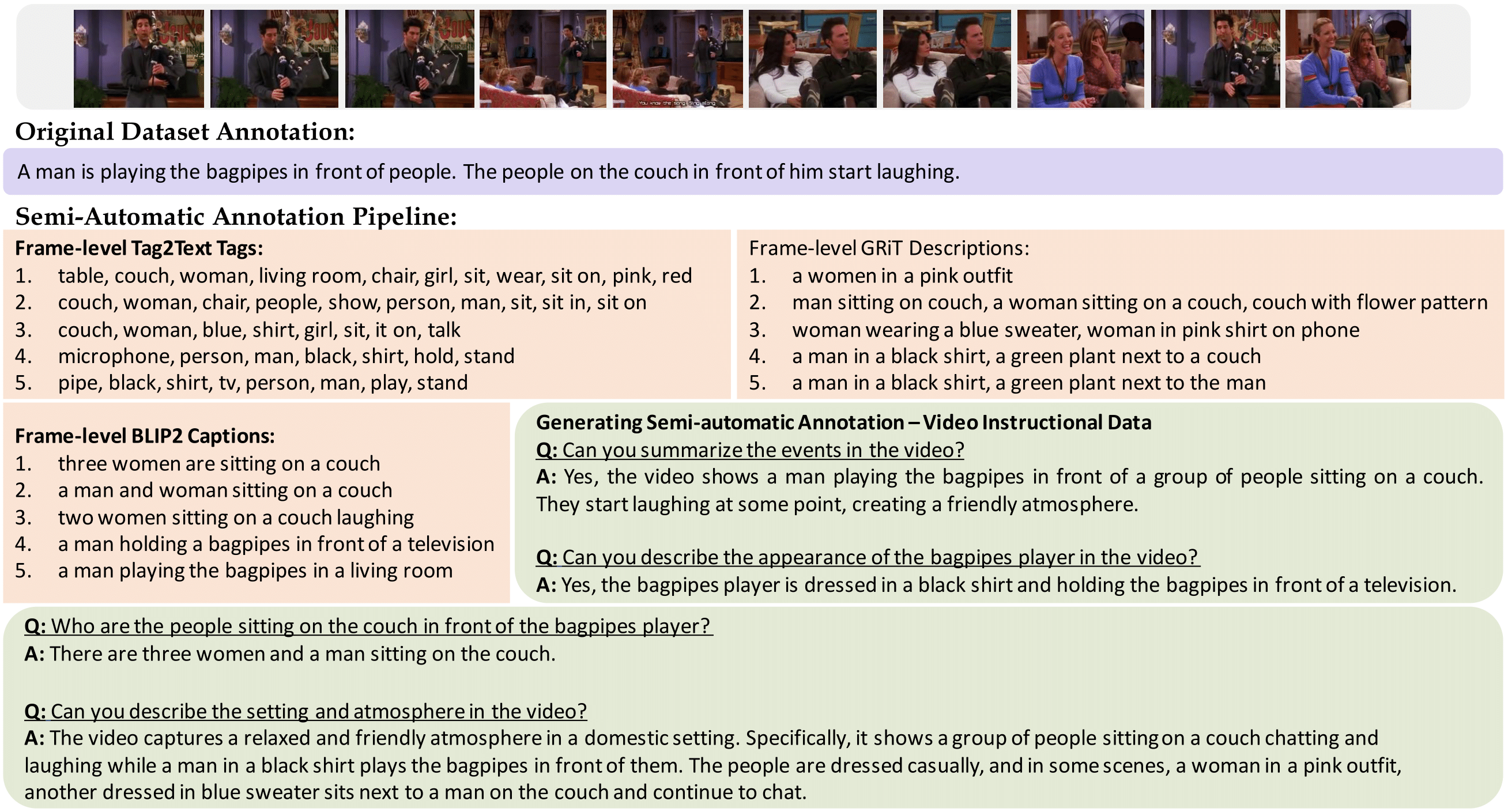}
  \caption{\textbf{Examples of generating instructional data using our proposed semi-automatic annotation pipeline.} We employ off-the-shelf dense prediction and captioning models to augment video descriptions. BLIP-v2~\cite{blip-2} generates frame-level captions, while GRIT~\cite{wu2022grit} is utilized for dense frame captions. Tag2Text~\cite{huang2023tag2text} generates tags for each key-frame, aiding in eliminating noise (e.g. the GRiT descriptions containing \textit{flower pattern} and \textit{on phone} would be discarded as there are no corresponding tags detected). Finally, we query GPT-3.5 with in-context examples to generate video-instructional data.}
  \label{fig:semi_automated-1}
\end{figure*}

\begin{figure*}[t]
  \centering
  \includegraphics[width=0.99\textwidth]{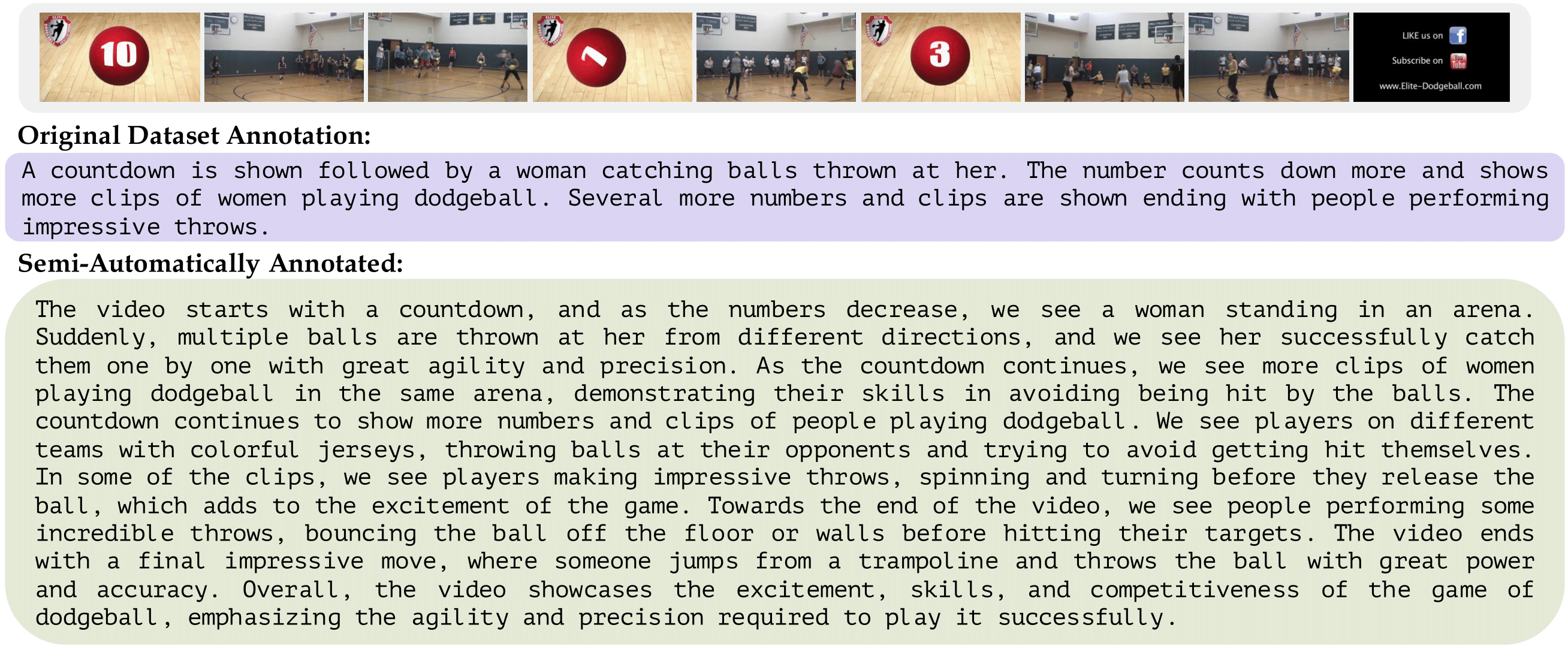}
  \caption{\textbf{Examples of data enrichment using our proposed semi-automatic annotation.} We employ off-the-shelf dense prediction and captioning models \cite{blip-2, wu2022grit, huang2023tag2text} to augment video descriptions. All additional context elements are combined with the video captions and undergo a GPT-assisted post-processing stage, generating the final detailed description.}
  \label{fig:semi_automated}
\end{figure*}

\subsection{Human-assisted Annotation}
In this process, we leverage datasets containing video-caption pairs and utilize the expertise of human annotators to enrich the original ground truth annotations. Specifically, we use a subset of ActivityNet-200~\cite{caba2015activitynet} which provides concise ground truth descriptions of various activities in distinct video segments.

The annotators further enrich the captions by adding comprehensive information about physical appearances and spatial and temporal localization, among other critical contextual details. Figure \ref{fig:human_assisted} shows an example of how a ground truth caption is enriched using human-assisted annotation.

\subsection{Semi-automatic Annotation Framework} In addition to the rich human-assisted annotations, we also harness the capabilities of advanced dense image vision-language models, developing a semi-automatic annotation framework. This approach is cost-effective and scalable, thereby increasing the quantity of data without substantially compromising the quality.

Similar to the human-assisted process, this framework also leverages datasets containing video-caption pairs. We enrich these datasets using contextual information drawn from off-the-shelf dense prediction and captioning image-based vision-language models. These models provide predictions that deliver additional contextual information, thereby enriching the video captions. We developed a comprehensive method that combines these predictions, and utilize specific models for the purpose of eliminating noisy or irrelevant context from the data. This ensures that the data maintains its accuracy and relevance.

Building on the use of off-the-shelf models, we apply pretrained models like BLIP-2~\cite{blip-2} and GRiT~\cite{wu2022grit} for key-frame analysis in the videos. The BLIP-2 image-captioning model generates frame-level captions, while the GRiT dense captioning model provides detailed captions for scene objects. Additionally, the pretrained Tag2Text~\cite{huang2023tag2text} model is used to generate tags for each key-frame of the video. Despite their utility, these models can introduce noise into the data.

To ensure high-quality data and mitigate noise, we implement three key steps. \emph{First,} we maintain a high prediction threshold for all off-the-shelf models to uphold accuracy. \emph{Second,} we employ a specialized filtering mechanism that removes any frame-level caption from BLIP-2 or GRiT not matching with the Tag2Text frame-level tags. This process involves extracting words from the frame-level captions that are within the predefined Tag2Text tags vocabulary and eliminates any captions that contain words not in the tags for a given frame. This strategy acts as an additional filtering layer and enriches the captions by integrating predictions from multiple models.

In the \emph{third} step, we merge frame-level captions and use the GPT-3.5 model to generate a singular, coherent video-level caption. This step augments the original ground truth caption with context from these models. We also direct GPT-3.5 to discard inconsistent information across frames, ensuring a precise, contextually rich video instruction dataset. Figure~\ref{fig:semi_automated-1},\ref{fig:semi_automated} illustrates how a ground truth caption is enriched using this process after all three refinement stages to generate instructional data and detailed descriptive caption. All of our designed prompts for in-context learning along with the curated dataset will be made publicly available.

\subsection{GPT-Assisted Postprocessing} 
Lastly, we implement a GPT-Assisted Post-processing mechanism that refines and optimizes the enriched annotations, in order to generate high-quality video instructional data. We prompt GPT-3.5 model to create question-answer pairs from the enriched and detailed captions that cover a wide variety of aspects using in-context learning. These aspects include detailed descriptions, summarizations, question-answer pairs, tasks that stimulate creativity or the generation of new ideas, and conversational tasks.

Each of these elements plays a crucial role in our data-centric approach. Our ultimate goal is to create a video-based conversation model that is accurate, capable of understanding video content from both spatial and temporal cues, and adept at engaging in conversations.

\section{Experiments}

\begin{table*}[t]
\centering
\renewcommand{\arraystretch}{1.0} 
\scalebox{0.85}{
\begin{tabular}{lcccc}
\toprule
\textbf{Evaluation Aspect} & \textbf{Video Chat} & \textbf{LLaMA Adapter} & \textbf{Video-LLaMA} & \textbf{Video-ChatGPT} \\ 
\midrule
\midrule
Correctness of Information & 2.23 & 2.03 & 1.96 & \textbf{2.40}\\
Detail Orientation & 2.50 & 2.32 & 2.18 & \textbf{2.52}\\
Contextual Understanding & 2.53 & 2.30 & 2.16 & \textbf{2.62}\\
Temporal Understanding & 1.94 & \textbf{1.98} & 1.82 & \textbf{1.98}\\
Consistency & 2.24 & 2.15 & 1.79 & \textbf{2.37}\\
\bottomrule
\end{tabular}}
\vspace{0.2em}
\caption{\textbf{Performance benchmarking of text generation models.} An in-depth comparative analysis of Video-ChatGPT and Video Chat~\cite{2023videochat} across five key evaluation aspects we propose in our benchmark. For a fair comparison, 7B variants are used for all the models. Video-ChatGPT shows competent performance across all key aspects.}
\label{table:1}
\end{table*}

\begin{table*}[ht]
\centering
\setlength{\tabcolsep}{8pt}
\renewcommand{\arraystretch}{1}
\scalebox{0.9}{
\begin{tabular}{l c c c c c c c c}
\hline
\textbf{Model} & \multicolumn{2}{c}{\textbf{MSVD-QA}} & \multicolumn{2}{c}{\textbf{MSRVTT-QA}} & \multicolumn{2}{c}{\textbf{TGIF-QA}} & \multicolumn{2}{c}{\textbf{Activity Net-QA}} \\
\cline{2-9}
 & \textbf{Accuracy} & \textbf{Score} & \textbf{Accuracy} & \textbf{Score} & \textbf{Accuracy} & \textbf{Score} & \textbf{Accuracy} & \textbf{Score} \\
\hline
\hline
FrozenBiLM & 32.2 & -- & 16.8 & -- & 41.0 & -- & 24.7 & -- \\
Video Chat & 56.3 & 2.8 & 45.0 & 2.5 & 34.4 & 2.3 & 26.5 & 2.2 \\
LLaMA Adapter & 54.9 & 3.1 & 43.8 & 2.7 & - & - & 34.2 & 2.7 \\
Video LLaMA & 51.6 & 2.5 & 29.6 & 1.8 & - & - & 12.4 & 1.1 \\
Video-ChatGPT & \textbf{64.9} & \textbf{3.3} & \textbf{49.3} & \textbf{2.8} & \textbf{51.4} & \textbf{3.0} & \textbf{35.2} & \textbf{2.8} \\
\hline
\end{tabular}}
\caption{\textbf{Zeroshot question-answering} comparison of Video-ChatGPT with other video generative models. For a fair comparison, 7B variants are used for all the models. Video-ChatGPT performs competitively across all datasets.}
\end{table*}

\subsection{Implementation Details}
We use LLaVA~\cite{liu2023llava} as our baseline model and finetune it on our 100K video instruction pairs. We only update the linear layer projecting the video features to the LLMs' input space, while the rest of the architecture is kept frozen. We finetune the model for 3 epochs using a learning rate of 2$e^{-5}$ and an overall batch size of 32. We use 7B parameter model in all the experiments and its training took around 3 hours on 8 A100 40GB GPUs. During inference, for memory efficiency, we load the models in FP16 mode.

In our semi-automatic annotation framework, we use Katna~\cite{Katna} to extract video key-frames. For off-the-shelf Tag2Text~\cite{huang2023tag2text} model, we use the Swin-B variant with an input size of 384$\times$384 and a confidence threshold of 0.7. For GRIT~\cite{wu2022grit}, we use ViT-B version with CenterNet2~\cite{zhou2021probablistic}.

\subsection{Quantitative evaluation}
\label{quantitative_eval}
In this section, we highlight a key contribution of our work: the quantitative evaluation of Video-ChatGPT using advanced metrics and comparative evaluations with existing state-of-the-art models. We conduct two types of quantitative evaluations: i) Video-based Generative Performance Benchmarking and ii) Zero-Shot Question-Answer Evaluation.

\noindent\textbf{Video-based Text Generation Performance Benchmarking: }
We introduce a benchmark to evaluate the text generation performance of video-based conversation models. To do this, we curate a test set based on the ActivityNet-200 dataset~\cite{caba2015activitynet}, featuring videos with rich, dense descriptive captions and associated question-answer pairs from human annotations. We also develop an evaluation pipeline using the GPT-3.5 model. This pipeline assesses various capabilities of the model and assigns a relative score to the generated predictions on a scale of 1-5, in the following five aspects:
\begin{enumerate}[label=(\roman*)] 
    \item \textit{Correctness of Information:} We verify the accuracy of the generated text, ensuring it aligns with the video content and does not misinterpret or misinform.
    \item \textit{Detail Orientation:} We evaluate the depth of the model's responses, looking for both completeness, meaning the model's response covers all major points from the video, and specificity, denoting the inclusion of specific details rather than just generic points in the model's response.
    \item \textit{Contextual Understanding:} We assess the model's understanding of the video's context, checking if its responses align with the overall context of the video content.
    \item \textit{Temporal Understanding:} We examine the model's grasp of the temporal sequence of events in the video when answering questions.
    \item \textit{Consistency:} We evaluate the model's consistency across different but similar questions or different sections of the video.
\end{enumerate}

We present the evaluation results of our proposed model, Video-ChatGPT, using the quantitative benchmarking framework in Table \ref{table:1}. The results reveal its competent performance across all key aspects compared with the recently introduced contemporary video conversation models, Video Chat~\cite{2023videochat}, LLaMA Adapter~\cite{gao2023llamaadapterv2} and Video-LLaMA~\cite{damonlpsg2023videollama}. Video-ChatGPT shows good performance, largely due to the instruction tuning we perform and its straightforward architecture that leverages LLMs with a pretrained visual encoder fine-tuned for video data. This provides it with the robust ability to generate contextually relevant, detailed, and temporally accurate text from video input.

\noindent\textbf{Zero-Shot Question-Answer Evaluation: } 
We conducted a comprehensive quantitative evaluation using several commonly used open-ended question-answer datasets: MSRVTT-QA~\cite{xu2017video}, MSVD-QA~\cite{xu2017video}, TGIF-QA FrameQA~\cite{jang2017tgif}, and ActivityNet-QA~\cite{yu2019activitynet}. These evaluations were carried out in a zero-shot manner, employing GPT-assisted evaluation to assess the model's capabilities. This evaluation process measures the accuracy of the model's generated predictions and assigns a relative score on a scale of 1-5.

To benchmark Video-ChatGPT, we compared its performance with other significant models, such as FrozenBiLM~\cite{yang2022zero} and the generative video model, Video Chat, LLaMA Adapter and Video-LLaMA. FrozenBiLM is a model that adapts frozen bidirectional language models pretrained on Web-scale text-only data to multi-modal inputs, showing promising results in zero-shot VideoQA settings. Despite the solid foundation established by these models, Video-ChatGPT consistently outperformed them, achieving state-of-the-art (SOTA) performance across all datasets. These results indicate Video-ChatGPT's ability to understand video content and generate accurate, contextually rich answers to questions.

\subsection{Ablations}

\noindent
\textbf{Impact of Semi-Automatic Annotations: } We train Video-ChatGPT on two subsets: one with human annotations (30\% of our data) and one with semi-automatic annotations (70\%). The results in Table.~\ref{ablation_1} indicate that training solely with human-annotated data or semi-automatically generated data yields good performance.  The overall performance when using only human-generated data is the lowest due to the limited number of labels (30\% of all data) available in this scenario. However, the optimal results are achieved when utilizing a combined dataset for training.

\begin{table}[h]
\centering
\label{table:video_chatgpt_comparison}
\resizebox{1.0\columnwidth}{!}{%
\begin{tabular}{lccc}
\toprule
\textbf{Metric} & \textbf{Human only} & \textbf{Automatic only} & \textbf{Combined} \\ \midrule
Correctness & 2.27 & 2.35 & 2.40 \\
Detail Orientation & 2.49 & 2.49 & 2.52 \\
Contextual Understanding & 2.50 & 2.56 & 2.62 \\
Temporal Understanding & 1.85 & 1.92 & 1.98 \\
Consistency & 2.21 & 2.38 & 2.37 \\
\midrule
Average & \textbf{2.28} & \textbf{2.34} & \textbf{2.38} \\
\bottomrule
\end{tabular}%
}
\caption{\small \textbf{Human Annotated vs Semi-automatically Annotated Data:} Training using both human annotated and semi-automatically annotated data achieves best performance.}
\label{ablation_1}
\end{table}

\noindent
\textbf{Quantitative Evaluation with GPT-3.5}:
Considering the limitations posed by the use of GPT-3.5, which is accessed via API and is not open-source, we perform evaluations using the open-source LLM, Vicuna-1.5 (13B)~\cite{vicuna2023}. 
The results in Table.~\ref{ablation_2} show similar trend in correctness, detail, contextual and temporal understanding and consistency compared with the initial GPT-3.5 evaluation.
This ensures our evaluation method remains accessible and replicable.

\begin{table}[h]
\centering
\label{table:video_chatgpt_evaluation}
\resizebox{1.0\columnwidth}{!}{%
\begin{tabular}{lccc}
\toprule
\textbf{Metric} & \textbf{Video Chat} & \textbf{Video-LLaMA} & \textbf{Video-ChatGPT} \\ \midrule
Correctness & 2.32 & 2.10 & 2.49 \\
Detail Orientation & 2.50 & 2.18 & 2.52 \\
Contextual Understanding & 2.76 & 2.41 & 2.85 \\
Temporal Understanding & 2.27 & 2.17 & 2.38 \\
Consistency & 2.95 & 2.67 & 3.09 \\ \bottomrule
\end{tabular}%
}
\caption{\small \textbf{Evaluation using Vicuna-1.5 (13B) Model:} We observe similar trend when evaluating using open-source Vicuna-1.5 (13B) model versus GPT-3.5-Turbo.}
\label{ablation_2}
\end{table}

\noindent
\textbf{Ensuring Automatic Annotation Pipeline Consistency:}
To ensure consistency between our automatic evaluation pipeline and human assessments, we conducted a blind test comparing QA pairs from human and semi-automatically annotated sources using 50 randomly sampled videos. A 52\% accuracy rate in distinguishing between the two demonstrated the reliability of our semi-automatic data, confirming that our quality control effectively aligns automatic evaluations with human judgment standards.

\section{Conclusion}
In this work, we presented Video-ChatGPT, a multimodal model that merges a pretrained visual encoder with a large language model (LLM) to enable video understanding and conversations based on videos. Video-ChatGPT leverages an adapter on top of pretrained LLM and vision backbones and is fine-tuned on video-instruction data to capture temporal dynamics and spatial consistency relationships in spatiotemporal sequences. A dataset of 100,000 video-instruction pairs is created to enhance Video-ChatGPT's video-specific understanding and conversation capabilities. The work also introduced a quantitative video conversation evaluation framework for benchmarking, evaluating models on a diverse set of capabilities including conventional video question answering as well as open-ended descriptions.

\section{Limitations}
While the model performs competitively in several scenarios, we note it finds it challenging to understand subtle temporal relationships in long videos ($>$ 2 min), which can compromise its predictive performance. Additionally, it has difficulty recognizing the details of small objects, often missing additional information embedded in these details.

\section{Potential Risks}
Video-ChatGPT, like any other AI model, must be handled with due caution to prevent misuse and to ensure it upholds the principles of fairness, transparency, and respect for user privacy. 

We made a concerted effort to minimize bias during the dataset creation phase for Video-ChatGPT. Despite these efforts, it is important to recognize the possibility of residual bias persisting. The use of our model should be mindful of these potential biases, which may subtly influence the model's understanding and response to visual content. We encourage all users to consider these limitations in their application of Video-ChatGPT and to strive for ethical and responsible use in all contexts.

\section{Use of Data and AI Assistant}
We curate our dataset based on a subset of the ActivityNet-200 dataset~\cite{caba2015activitynet}, distributed under MIT LICENSE, available for use in research. 
Further, the use of GPT models abides by~\cite{openai}. 
Respecting source license information, we will release all datasets created in this work under MIT LICENSE.

\section{Human Annotations}
The semi-automatic dataset curation involves human annotation. Annotators are provided with concise video caption ground truths. Specific instructions are given to enrich the caption with comprehensive descriptions of the video content, with specific attention to temporal and spatial details. They are given specific instructions to neutralize the tone and biases during the correction process.

\section{Qualitative Evaluation}
\begin{figure*}[h]
  \centering
  \includegraphics[width=0.75\textwidth]{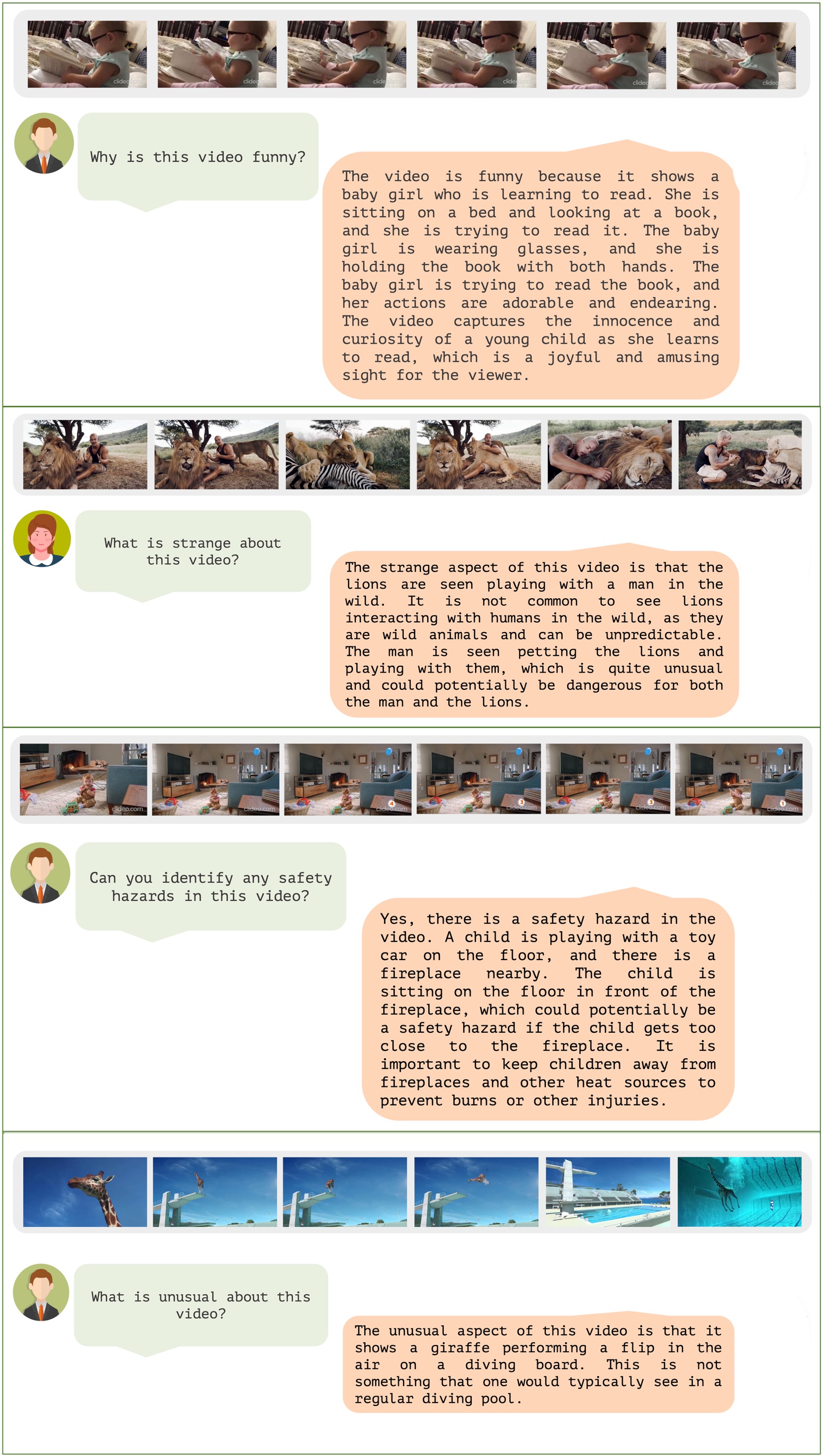}
  \caption{\textbf{Video Reasoning Task.} This figure illustrates an example from Video-ChatGPT's demonstration showcasing its performance in video reasoning tasks.}
  \label{fig:video_reasoning}
\end{figure*}

\begin{figure*}[h]
  \centering
  \includegraphics[width=0.75\textwidth]{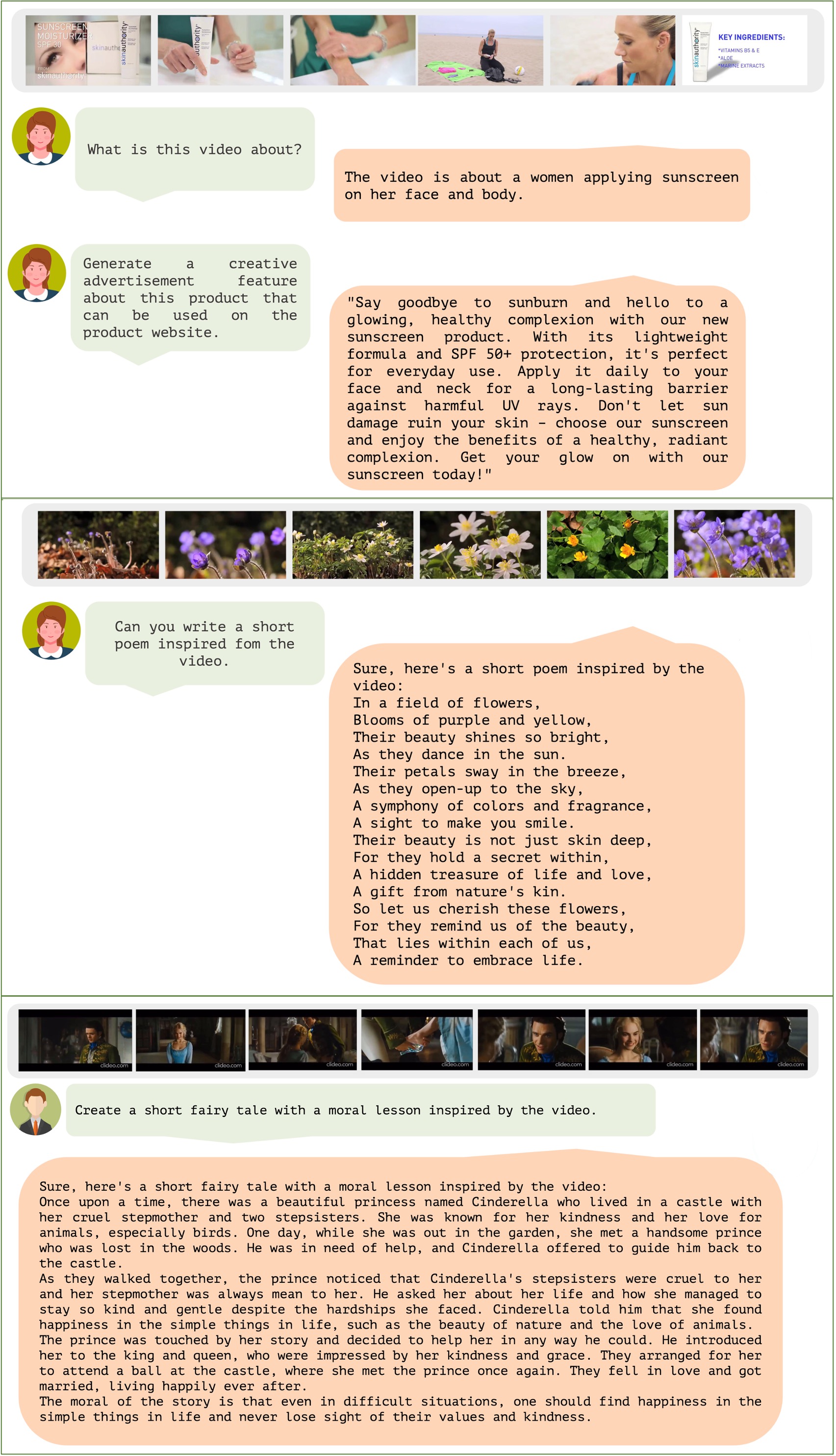}
  \caption{\textbf{Creative and generative tasks}. Illustrative examples from Video-ChatGPT's demonstration highlight its performance in video-based creative and generative tasks, such as crafting a story, poem, or advertisement.}
  \label{fig:creative_generative}
\end{figure*}

\begin{figure*}[h]
  \centering
  \includegraphics[width=0.725\textwidth]{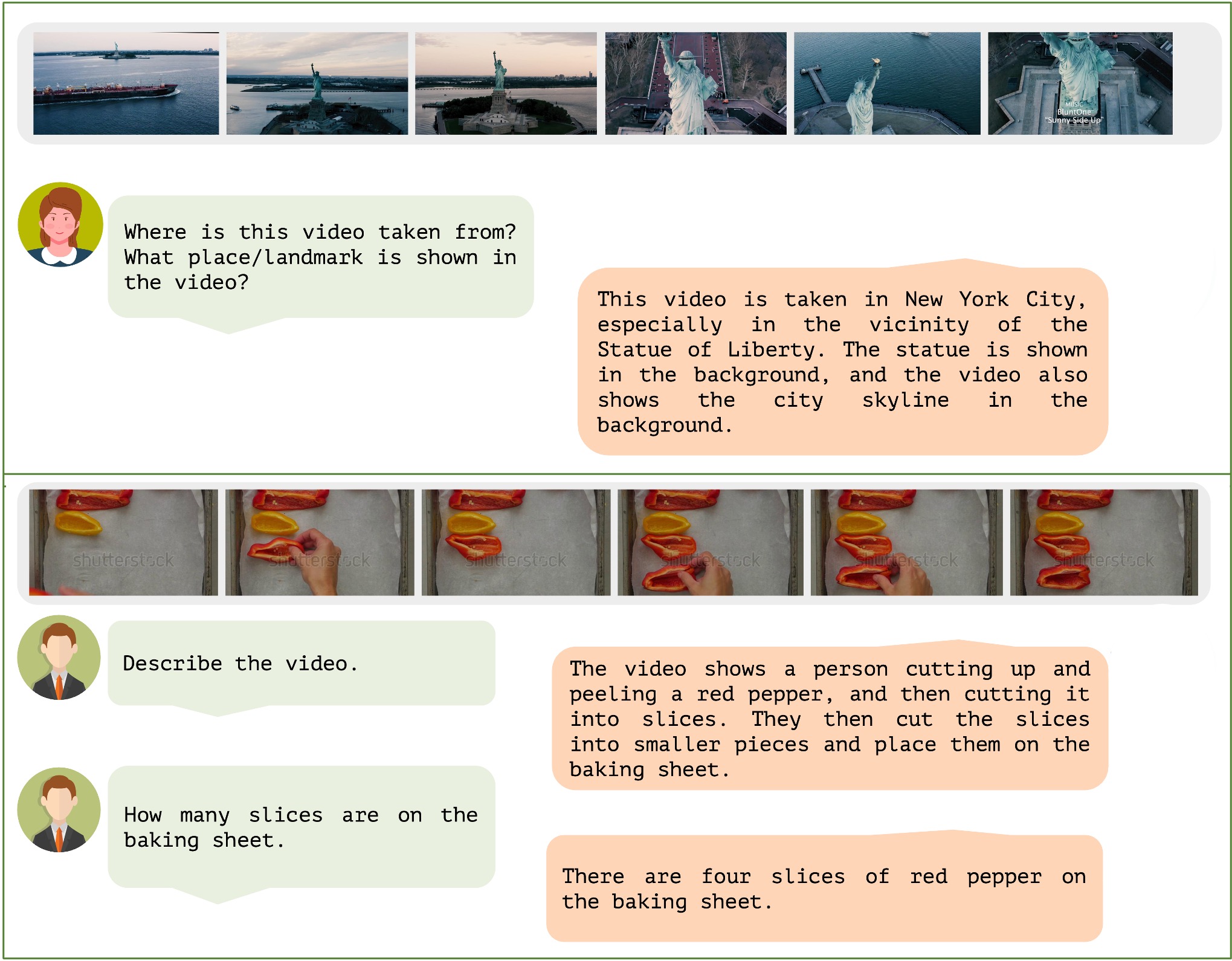}
  \caption{\textbf{Spatial understanding tasks.} The figure depicts examples from Video-ChatGPT's demonstration, emphasizing its capability in video-based spatial understanding tasks, including identifying renowned locations or counting the number of objects in a scene.}
  \label{fig:spatial_understanding}
\end{figure*}

\begin{figure*}[hb]
  \centering
  \includegraphics[width=0.725\textwidth]{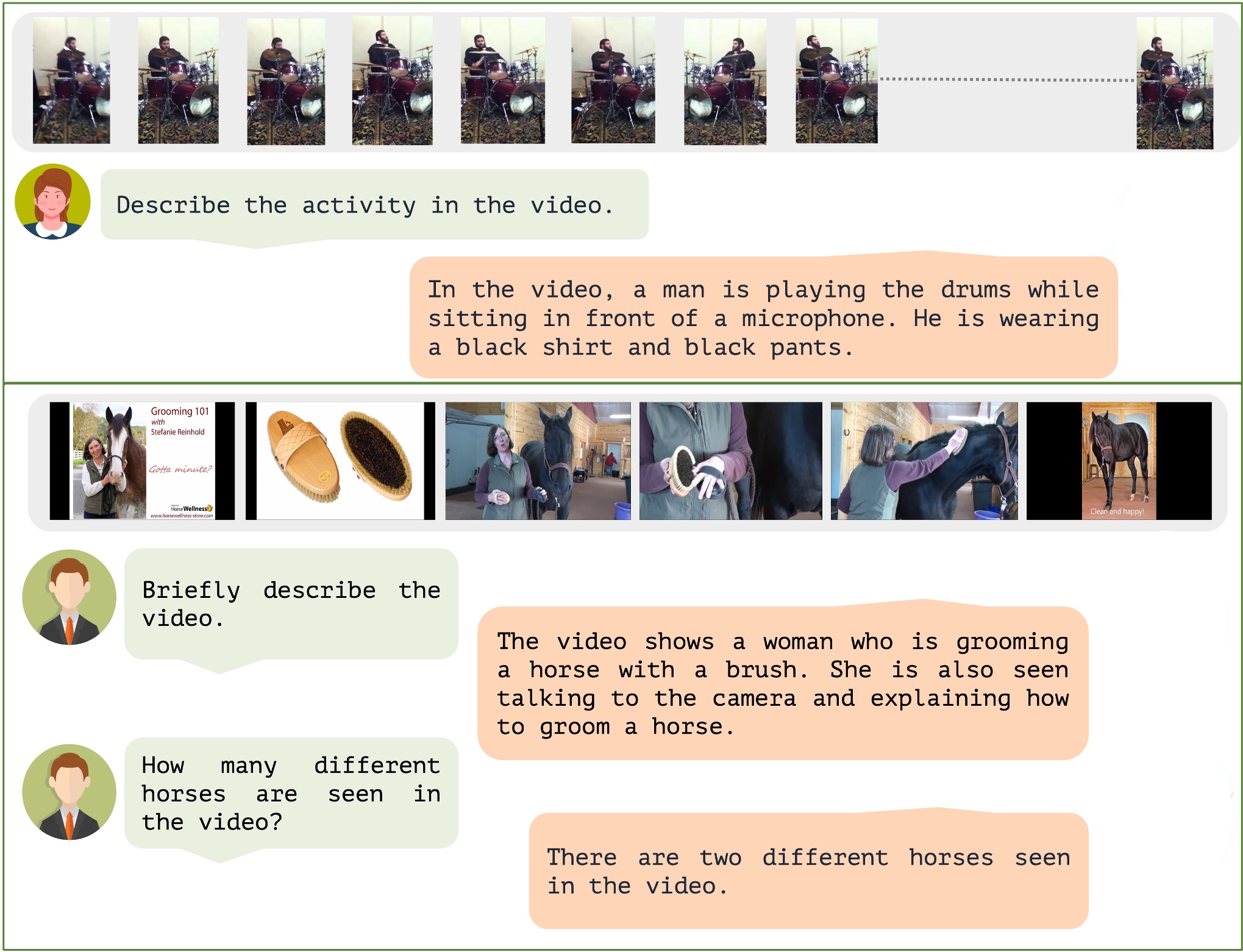}
  \caption{\textbf{Actiong Recognition Task.} This figure illustrates examples from Video-ChatGPT's demonstration showcasing its performance in video action recognition tasks such as playing drums and grooming horse.}
  \label{fig:action_recognition}
\end{figure*}

\begin{figure*}[ht]
  \centering
  \includegraphics[width=0.75\textwidth]{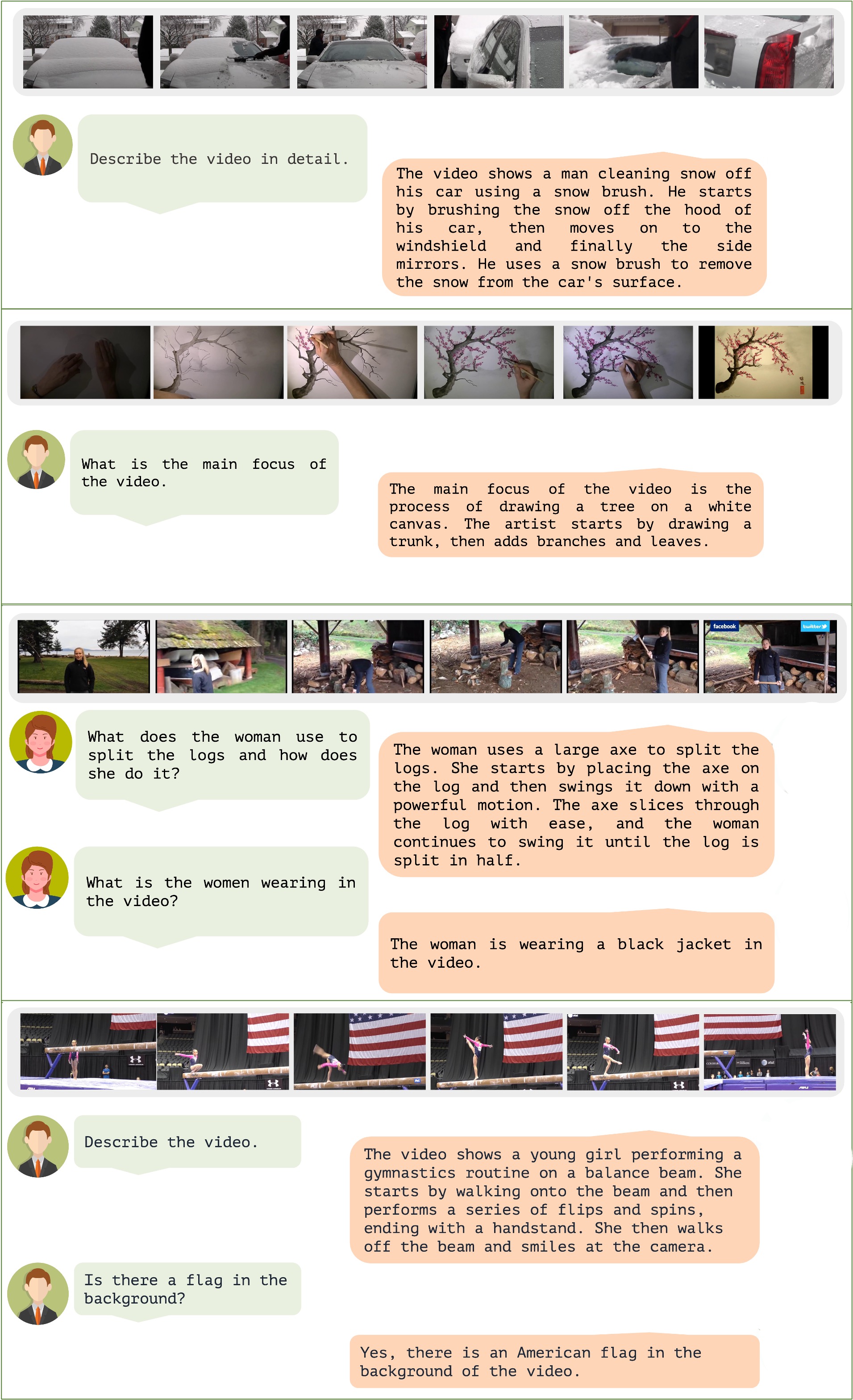}
  \caption{\textbf{Video Understanding and Conversation Tasks.} This figure illustrates examples from Video-ChatGPT's demonstration showcasing its performance in video understanding and conversation tasks.}
  \label{fig:video_understanding_conversation}
\end{figure*}

\begin{figure*}[ht]
  \centering
  \includegraphics[width=0.75\textwidth]{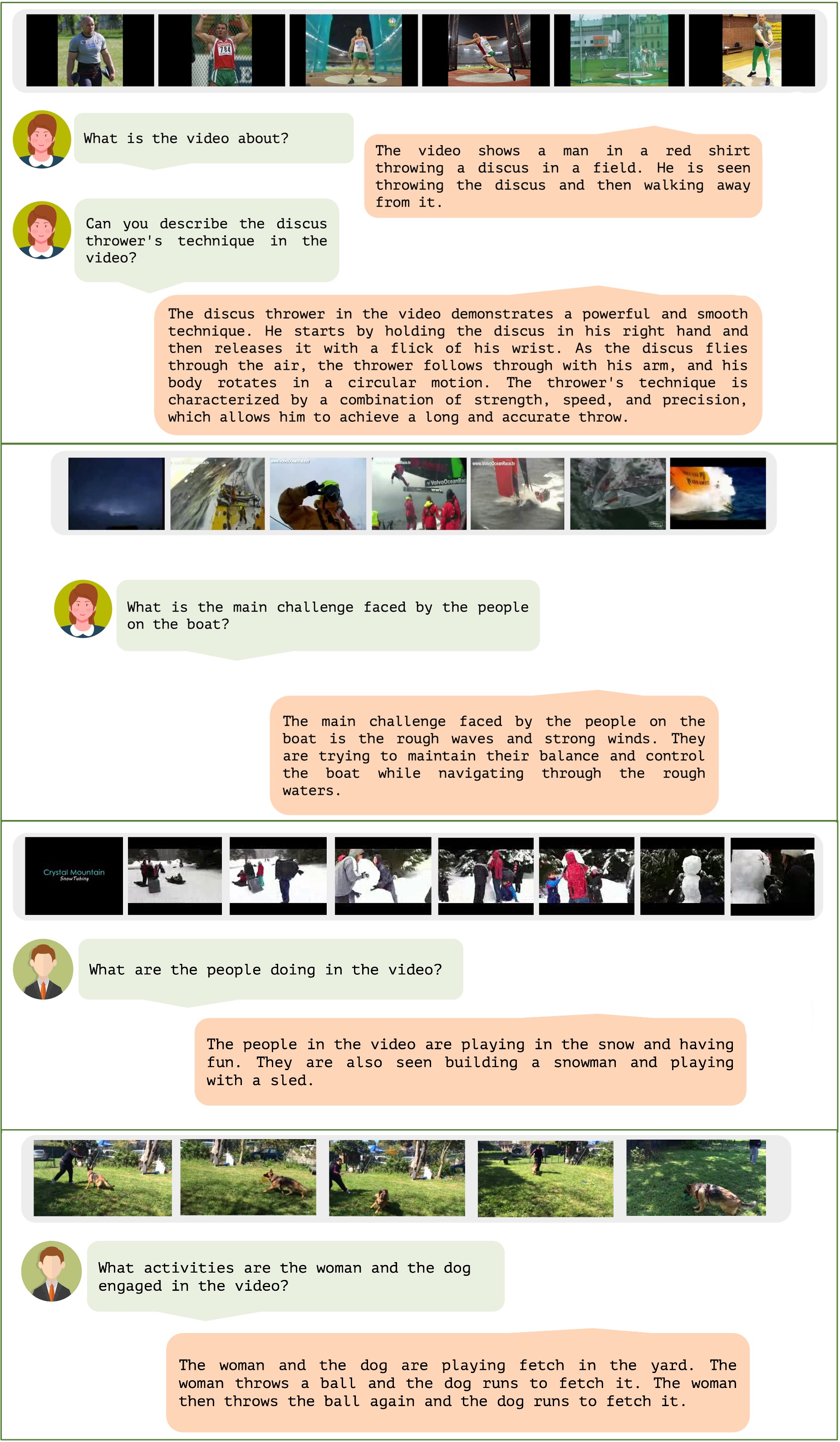}
  \caption{\textbf{Question-Answering Task.} The figure depicts examples Video-ChatGPT's demonstration showcasing its performance in question-answering tasks.}
  \label{fig:question_answer}
\end{figure*}

\begin{figure*}[ht]
  \centering
  \includegraphics[width=0.72\textwidth]{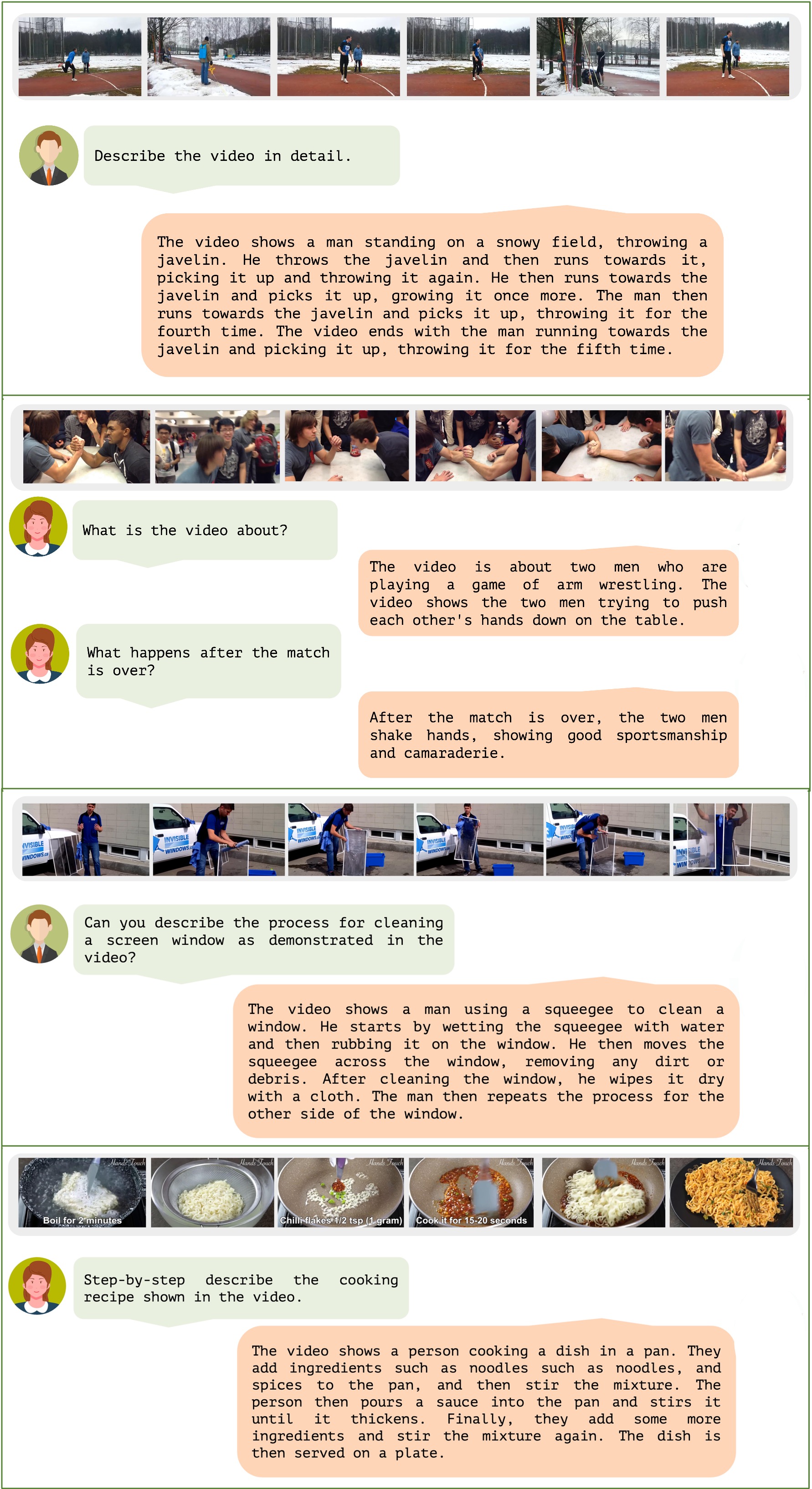}
  \caption{\textbf{Temporal Understanding Task.} The figure provides examples from Video-ChatGPT's demonstration, highlighting its performance in temporal understanding tasks, particularly in comprehending sequences of events.}
  \label{fig:temporal_understanding}
\end{figure*}

We performed an extensive evaluation of our model on a variety of open-ended video question-answering tasks, utilizing diverse videos sourced from ActivityNet and YouTube. The evaluation tasks included video reasoning (\Cref{fig:video_reasoning}), creative and generative tasks (see~\Cref{fig:creative_generative}), spatial understanding (\Cref{fig:spatial_understanding}), action recognition (\Cref{fig:action_recognition}), video conversation (\Cref{fig:video_understanding_conversation}), question answering (\Cref{fig:question_answer}) and temporal understanding (\Cref{fig:temporal_understanding}). Our model demonstrates proficiency in comprehending the content of the videos and generating accurate responses across multiple video-based tasks. Our model can effectively understand the visual information present in the videos and provide precise answers (see~\Cref{fig:video_reasoning,fig:creative_generative,fig:spatial_understanding,fig:action_recognition,fig:video_understanding_conversation,fig:question_answer,fig:temporal_understanding}).

\bibliography{acl_latex}

\clearpage
\appendix

\end{document}